\documentclass[11pt]{article}
\usepackage{geometry}
\usepackage{coling2020}
\usepackage{times}
\usepackage{url}
\usepackage{latexsym}
\usepackage{microtype}
\usepackage{latexsym}
\usepackage{mathtools}
\usepackage{amsmath} 
\usepackage{subcaption}

\colingfinalcopy 

\author{Mayank Raj\thanks{\textsuperscript{*}Equal Contribution} \\  {\bf Ankita Gupta} \\\And Ajay Jaiswal\footnotemark[1] \\ {\bf  Sudeep Kumar Sahoo } \\\And   Rohit R.R  \\ {\bf Vertika Srivastava } \\ \AND Yeon Hyang Kim \\ \AND
    Samsung R\&D Institute India, Bangalore\\
    \tt \{ mayank.raj, ajay.jaiswal, rohit.r.r, \\ \tt gupta.ankita, sudeep.sahoo, v.srivastava, purine.kim\}@samsung.com \\}

\title{Solomon at SemEval-2020 Task 11: Ensemble Architecture for Fine-Tuned Propaganda Detection in News Articles}

\date{}

\begin{document}
\maketitle
\begin{abstract}
  This paper describes our system (Solomon) details and results of participation in the SemEval 2020 Task 11  "Detection of Propaganda Techniques in News Articles"\cite{DaSanMartinoSemeval20task11}. We participated in Task "Technique Classification" (TC) which is a multi-class classification task. To address the TC task, we used RoBERTa based transformer architecture for fine-tuning on the propaganda dataset. The predictions of RoBERTa were further fine-tuned by class-dependent-minority-class classifiers. A special classifier, which employs dynamically adapted Least Common Sub-sequence algorithm, is used to adapt to the intricacies of repetition class. Compared to the other participating systems, our submission is ranked 4th on the leaderboard.
\end{abstract}
\blfootnote{
This work is licensed under a Creative Commons Attribution 4.0 International Licence.License details: http://creativecommons.org/licenses/by/4.0/.
}
\label{intro}
\section{Introduction}
In today's digital world, information is being shared more rapidly across boundaries owing to the rampant use of social media platforms. While these platforms provide a forum for people to exchange their views with ease, they also serve as a medium for spreading misinformation in the absence of proper quality control mechanisms. Two paradigms of misinformation are often reported, fake news and propaganda. While fake news corresponds to factually incorrect information, propaganda is more nuanced. It aims at advancing a specific agenda using psychological and rhetorical techniques (e.g use of biased or loaded language, repetition etc) and may not be always factually wrong. 

Prior efforts have been made to segregate propagandist content \cite{rashkin2017truth}, \cite{barron2019proppy}, \cite{habernal2017argotario} from non-propagandist. These methods use distant supervision based techniques to obtain document level labels, often labelling all articles from a propagandist news outlet as propaganda. However, such techniques introduce inevitable noise \cite{horne2018sampling}.
To this end, \cite{da2019fine} have introduced a propaganda dataset where they label propaganda techniques at fragment level, thus providing finer control and explainability. For example, following is an instance of propaganda with type ``Slogan". 

\begin{equation*}
    Trump\;\;tweeted,\;\;``\underbrace{make\;America\;great \;again\;!}_{Slogan}"
\end{equation*}


In this work we address the "Technique Classification" (TC) task whose objective is to identify the propaganda technique employed in a fragment of text surrounded by its context. In the above example, ``make America great again" is the fragment in which \textit{slogan} technique has been employed. It is surrounded by its context ``Trump tweeted". The problem is modelled as a 14-class classification task. Our contributions in this work are summarized as follows : 
\begin{itemize}
    \item  We employ transfer learning by fine-tuning Transformer based language model, RoBERTa~\cite{liu2019roberta} on the propaganda dataset.
    \item We address the issue of minority class classification by designing ensemble of \textit{one-versus-one (OVO)} classifiers, that vote for the presence/absence of the specific minority class.
    \item We handle the intricacies in the \textit{Repetition} class by employing a novel algorithm based on dynamic least common sub-sequence approach.
\end{itemize}

The remainder of this paper is organized as follows: Section 2 describes the system details for identification of propaganda technique employed in a sentence. Section 3 describes our experiments and the evaluation results. Finally, we conclude in Section 4.

\section{System Description}

\begin{figure}
    \includegraphics[width=\linewidth]{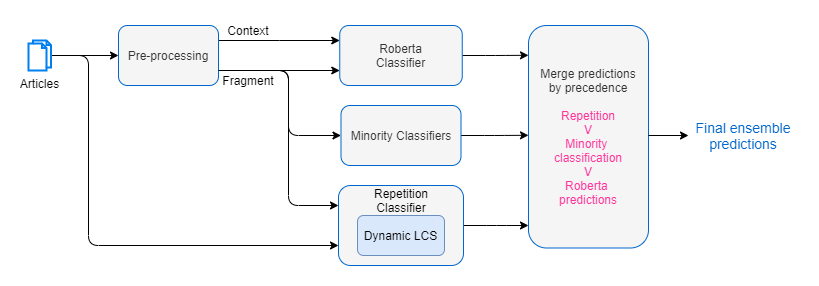}
    \caption{System Architecture with Precedence Order of Components}\label{fig:arch}
 \end{figure}
 
 \subsection{System Architecture}
Our system comprises of three major components as depicted in Fig~\ref{fig:arch}. The first component is fine-tuned RoBERTa that gives predictions for 14 classes. The predictions of RoBERTa are adjusted based on the output of Minority Classifiers followed by the Repetition classifier. If the Minority and Repetition classifier refrain from making a prediction, we retain the predictions of RoBERTa. The order of precedence among these three components is depicted in the architecture diagram.
\subsection{Fine Tuned RoBERTa}
Our first system component comprises of fine-tuning (end-to-end) a pre-trained language model on the downstream task of propaganda technique classification. We leverage RoBERTa, a transformer architecture based language
model for fine-tuning. RoBERTa uses BERT’s~\cite{devlin2018bert} masked language modelling strategy with modified hyper-parameters and is trained for longer period of time on substantially larger dataset (10 times more in size) in comparison to BERT. 

In order to fine-tune RoBERTa for 14-class TC task, we modify the last layer, comprising of 14 hidden nodes with softmax activation. This layer is specifically trained only on the downstream TC task, whereas all other layers of RoBERTa are fine-tuned.

We experimented with three methods of input, i.e., \textit{`fragment only'}, \textit{`context only'} and \textit{`fragment and context both'}. While the fragments are mainly decisive in classification, they face the issue of shorter length and missing background information, hence the reason for sub-optimal performance of \textit{`fragment only'} approach. Using the \textit{`context only'} approach, led to overshadowing the emphasis on the fragment embedded in a long surrounding text. Thus, the combined approach of using both fragment and context worked best as illustrated in our experiments section. 

We input both fragment and context as an input to RoBERTa by concatenating them together along with the separator to mark the beginning and end of each sequence. This is in similar lines to how RoBERTa takes input for a pair of sentence for tasks such as QA, NLI etc ~\cite{liu2019roberta}. Using fragment and context as a pair of input sentences to RoBERTa, helps us leverage bidirectional cross attention between the two sentences. The predictions of RoBERTa are then adjusted further based on the Minority and Repetition Classifiers as described next.  

\begin{figure}
\begin{subfigure}{.5\textwidth}
  \centering
  \includegraphics[width=.8\linewidth, height=4.2cm]{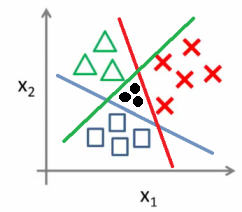}
  \caption{2a}
  \label{fig:ooa}
\end{subfigure}%
\begin{subfigure}{.5\textwidth}
  \centering
  \includegraphics[width=\linewidth, height=4.2cm]{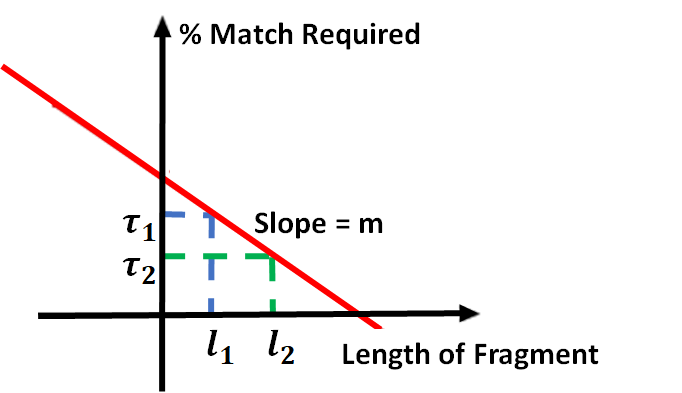}
  \caption{2b}
  \label{fig:sfig2}
\end{subfigure}
\caption{(a) : Ensemble of One-Versus-One Classifiers Segregating Minority Class.   (b) : Dynamic-LCS depicting the variation between $\%$ match required and length of fragment.}
\label{fig:dynamic-lcs}
\end{figure}

\subsection{Minority Classifiers}
The dataset has severe class imbalance owing to the varied frequency of occurrence of the different types of propaganda techniques in real life. For few minority classes, the number of positively labelled data samples are extremely small in comparison to the other classes. We tackle each one of these minority classes namely, \textit{``Bandwagon,Reductio ad Hitlerum"}, \textit{``Appeal to Authority"}, \textit{``Black and White Fallacy"}, \textit{``Whataboutism,StrawMen,RedHerring"}, and \textit{``Thought Terminating Cliche"} by training 5 separate hierarchical classifiers. 

We denote these 5 classifiers as \textit{level-1} classifier in the hierarchy. Each one of these \textit{level-1} classifier is further composed of an ensemble of \textit{n-1} (13 in our case) one-versus-one linear classifiers, denoted as \textit{level-2} classifiers. We then take vote of all these n-1 classifiers and aggregate their predictions to obtain the final prediction confidence of a \textit{level-1} classifier, denoting its confidence in predicting a corresponding minority class. If the confidence is above a threshold, we treat it as a positive example of that minority class. Rationale behind using such an approach is depicted in Fig~\ref{fig:ooa}, where learning three OVO classifier can clearly help segregate the minority class (black) from others. Also, this ensemble based classifier essentially ensures that each one of the \textit{n-1} classifiers should, with very high confidence, predict the positive class.  This helps us in classifying these minority classes within the data restricted settings and at the same time overcome the issue of possible over-fitting in such scenarios. We verify this intuitive explanation in our experiments.

A \textit{level-2} classifier is a simple linear classifier and hence lends itself reduced time and computational complexity as an added benefit. The final predictions of the \textit{level-1} minority class classifiers are used to over-rule the predictions of RoBERTa. In cases where the prediction confidence is less than the threshold, predictions of RoBERTa take precedence.

\subsection{Repetition}
\label{sec:minority}
Repetition is another class that we handle separately based on our analysis of the data. Repetition class, by definition, refers to repeating the same phrase over and over again across the message to enhance its impact on the audience. We formulate this problem as detecting the presence of dynamic least common sub-sequence (dynamic-LCS) between the fragment and the context. We avoid using exact match as fragments are repeated across the message with some minor modifications. For instance, consider the fragment \textit{"ammunition was purchased under someone else's name"}. This fragment is repeated in various forms across the message, such as \textit{"ammo was bought under someone else's name"}.

In order to measure the presence of LCS, and hence repetition, we detect if the $\%$ match between fragment and context is greater than a threshold ($\tau$). This threshold is dynamically adapted based on the length of the fragment as shown in Fig~\ref{fig:dynamic-lcs}. If the length of the fragment is small, high LCS is desired for a reliable indication of sufficient match and vice-versa. We model this by tuning the slope parameter (on dev-set) of the curve between length of the fragment and desired match threshold. Mathematically, $\tau$ is defined as below.
\begin{equation}
\begin{split}
\tau = 100 - (m) * l 
\end{split}\label{eq:label2}
\end{equation}

For a given slope of m=0.2, a string of length 100 will need 80$\%$ match. This implies that at-least 8 characters out of 10 should match for a string length of 10. 


\section{Experiments and Analysis}
 \begin{figure}
 \centering
    \includegraphics[width=10cm, height = 5cm]{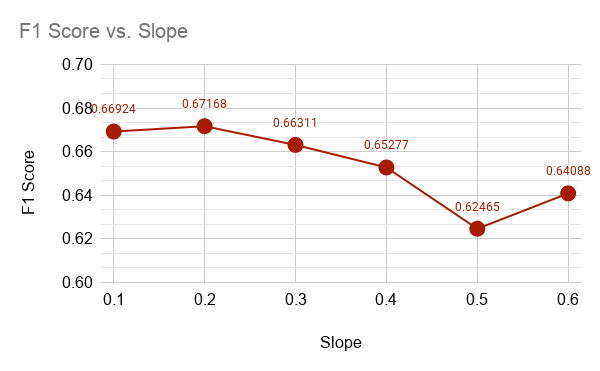}
    \caption{Variation of F1 score with respect to slope hyperparameter.}\label{fig:f1_slope}
 \end{figure}

 \begin{figure}
    \centering
    \includegraphics[width=10cm, height=5cm]{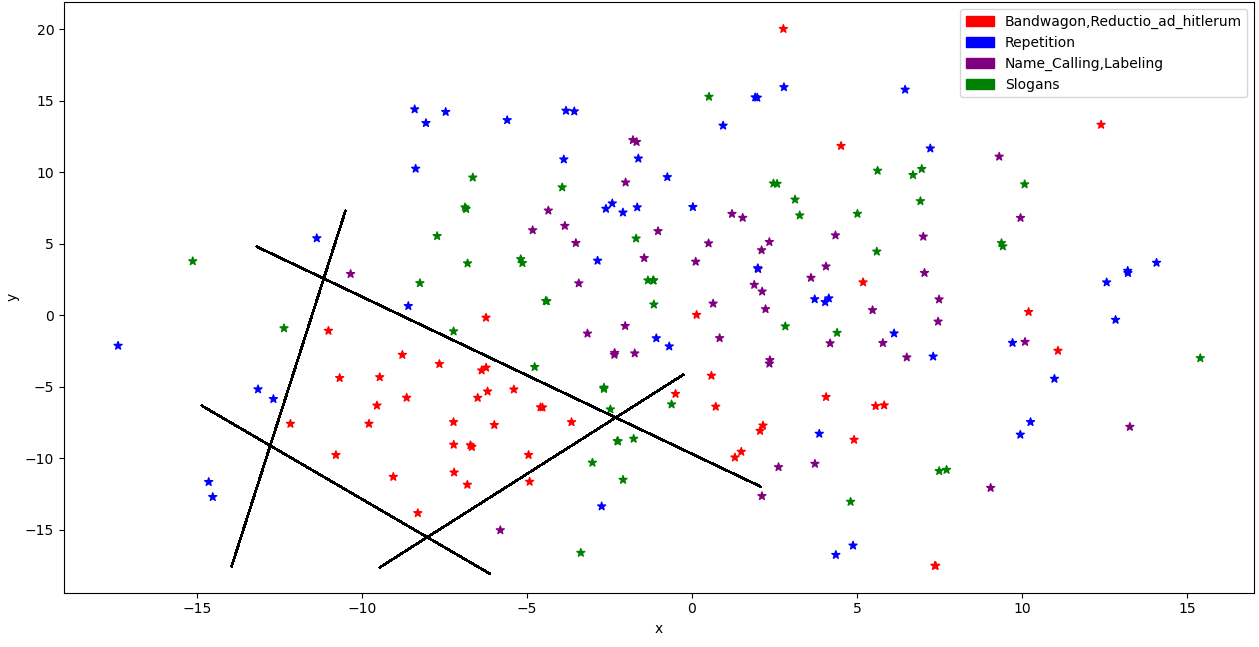}
    \caption{t-SNE plot of RoBERTa embedding}\label{fig:tsne}
 \end{figure}
 
\subsection{Analysis of fine-tuned RoBERTa}
Table~\ref{table:Roberta} mentions our experimental results obtained for various models. As a baseline, we also experimented with an SVM classifier (linear and RBF kernel) trained on pre-trained BERT embeddings of propagandist fragment. Also, for brevity we also checked the performance of fine-tuned BERT which was trained end-to-end similarly to how we trained RoBERTa. Our Fine-tuned RoBERTa gave the best performance with an an F1 score of 0.64, being pre-trained on remarkable larger dataset in comparison to fine-tuned BERT. Further, adapting the predictions of RoBERTa based on minority classifiers and Repetition classifier led to F1 scores of 0.65 and 0.67 respectively. 

We used uncased base model of BERT with batch size of 16, maximum sequence length of 512, weight decay of 0.01 and trained using adam optimizer with a learning rate of 3e-05 for 5 epochs.For RoBERTa, we used uncased large model with a batch size of 8, maximum sequence length of 512 and weight decay of 0.1. It was trained using Adam Optimizer with learning rate of 1e-05 for 10 epochs.

\subsection{Analysis of Minority Classifiers and Repetition Classifier}
As depicted in Table~\ref{table:Roberta}, predictions of fine-tuned RoBERTa perform well on all the majority classes. However, it exhibits poor F1 score for minority classes. We believe that high class imbalance in the dataset resulted in such an observation and motivated us to look closely at data distribution across classes. To illustrate our intuition introduced in Section~\ref{sec:minority}, we projected the sentence embeddings (RoBERTa) of minority samples onto a 2D subspace using t-Distributed Stochastic Neighbor Embedding (t-SNE) plot. Fig~\ref{fig:tsne} depicts how the data points corresponding to the minority class ``\textit{Bandwagon, Reductio\_ad\_hitlerum}" are arranged relative to other majority class samples. It can be observed that the minority samples in this case can be segregated well by learning an ensemble of linear classifiers, thus reinforcing our approach.

In experiments, we use n\_components=2, perplexity=40, n\_iter=300 while plotting t-sne plots. We also employ oversampling with replacement while training OVO \textit{level-2} classifiers. We kept the threshold for aggregated confidence of all \textit{level-2} classifiers as 0.95, above which a \textit{level-1} classifier will vote in favour of a minority class. As presented in Table~\ref{table:Roberta} Minority classifiers led to a $2\%$ boost in the F1 score. Specifically, the F1 score of \textit{Bandwagon, Reductio\_ad\_hitlerum} increased from 0.0 to 0.89 on the dev set.



Further improvement in F1 score can be observed due to Repetition Classifier in Table~\ref{table:Roberta}. We also measure how the system performance varies with the change in hyper-parameter slope ``m". Fig~\ref{fig:f1_slope} depicts that slope=0.2 gives best performance with highest F1-score of 0.67.

\begin{table}
    \centering
    \begin{tabular}{|l|l|}
      \hline
      \textbf{Models} & \textbf{F1 Score}\\
      \hline
      Official Baseline (Logistic Regression + Sentence length) & 0.25196\\
      \hline
      Pre-trained BERT embedding + SVM (Linear Kernel)  & 0.44120\\
      \hline
      Pre-trained BERT embedding + SVM (RBF Kernel)  & 0.48636\\
      \hline
      Fine-tuned BERT & 0.59548 \\
      \hline
      Fine-tuned Roberta & 0.63891 \\
      \hline
      Roberta + Minority Classifier & 0.65744\\
      \hline
      Roberta + Minority Classifier +  Repetition Classifier  &  0.67168\\
      \hline
    \end{tabular}
    \caption{Experiment results of different model for the 14 class propaganda detection}
    \label{table:Roberta}
\end{table}

\begin{table}[t!]
\centering
\begin{tabular}{|c|c|}
\hline \bf Techniques & \bf RoBERTa F1\\ \hline
Loaded Language & 0.78655 \\
Name Calling, Labeling & 0.74176\\
Repetition & 0.23015\\
Flag Waving & 0.74359\\
Exaggeration, Minimisation & 0.50746\\
Doubt & 0.61224\\
Slogans & 0.71605\\
Appeal to fear-prejudice & 0.41509\\
Causal Oversimplification & 0.31250\\
\hline
Appeal to Authority & 0.02809\\
Black \& White Fallacy & 0.01683\\
Whataboutism, StrawMen, RedHerring & 0.11432\\
Thought-terminating-Cliches & 0.27838\\
Bandwagon,Reductio-Hitlerum & 0.00000\\
\hline
\textbf{Total F1} & \textbf{0.63891} \\
\hline
\end{tabular}
\caption{\label{tab:f1} F1 Score of 14 classes with Fine-tuned RoBERTa Model. }
\end{table}


\section{Conclusion}
In this paper, we examine capability of the transformer based pre-trained language model, RoBERTa. We illustrated that Fine-tuned RoBERTa performed better on minority classes compared to BERT. We also introduced novel approach of handling the minority classes using an ensemble of one-vs-one simple classifiers. Furthermore, we handle the Repetition class separately using Dynamic LCS algorithm. Experiments show the improvement in F1 score when fine-tuned RoBERTa predictions are augmented with minority class classifiers and repetition classifier.

\bibliographystyle{coling}
\bibliography{semeval2020}

\end{document}